\documentclass[10pt,twocolumn,letterpaper]{article}

\usepackage{cvpr}
\usepackage{times}
\usepackage{epsfig}
\usepackage{graphicx}
\usepackage{amsmath}
\usepackage{amssymb}
\usepackage{subcaption}

\usepackage[pagebackref=true,breaklinks=true,letterpaper=true,colorlinks,bookmarks=false]{hyperref}

\cvprfinalcopy 

\def\cvprPaperID{****} 

\begin{document}

\title{SuperCaptioning: Image Captioning Using Two-dimensional Word Embedding}

\author{Baohua Sun,  ~Lin Yang, ~Michael Lin,  ~Charles Young, ~Patrick Dong, ~Wenhan Zhang, ~Jason Dong\\
Gyrfalcon Technology Inc.\\
1900 McCarthy Blvd. Milpitas, CA 95035\\
{\tt\small \{baohua.sun,lin.yang\}@gyrfalcontech.com}
}

\maketitle

\begin{abstract}
   Language and vision are processed as two different modal in current work for image captioning.
However, recent work on Super Characters method shows the effectiveness of two-dimensional word embedding, which converts text classification problem into image classification problem. In this paper, we propose the SuperCaptioning method, which borrows the idea of two-dimensional word embedding from Super Characters method, and processes the information of language and vision together in one single CNN model. The experimental results on Flickr30k data shows the proposed method gives high quality image captions. An interactive demo is ready to show at the workshop.

\end{abstract}

\section{Introduction}
Image captioning outputs a sentence related to the input image. Current methods process
the image and text separately~\cite{karpathy2015deep,vinyals2016show,you2016image,yao2017boosting,lu2017knowing,rennie2017self,anderson2018bottom,hossain2019comprehensive}. Generally, the image is processed by a CNN model to extract the image feature, and the raw text passes through embedding layer to convert into one-dimensional word-embedding vectors, e.g. a 300x1 dimensional vector. And then the extracted image feature and the word embedding vectors will be fed into another network, such as RNN, LSTM, or GRU model, to predict the next word in the image caption sequentially.

Super Characters method~\cite{sun2018superShort} is originally designed for text classification tasks. It has achieved state-of-the-art results on benchmark datasets for multiple languages, including English, Chinese, Japanese, and Korean. It is a two-step method. In the first step, the text characters are printed on a blank image, and the generated image is called Super Characters image. In the second step, the Super Characters image is fed into a CNN model for classification. The CNN model is fine-tuned from pre-trained ImageNet model. The extensions of Super Characters method~\cite{sun2019squared,sun2019supertml,sun2019superchat} also prove the effectiveness of two-dimensional embedding on different tasks.

\begin{figure}[t!]
\centering
\begin{subfigure}[t]{3.6cm}
\includegraphics[width=1\linewidth]{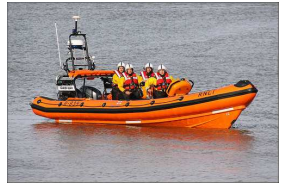}
\caption{``Four men in life jackets are riding in a bright orange boat".}\label{FourOrange}
\end{subfigure}\qquad
\begin{subfigure}[t]{3.6cm}
\includegraphics[width=1\linewidth]{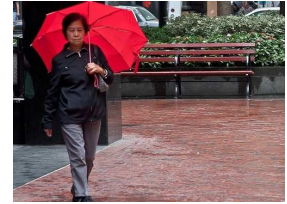}
\caption{``A woman in a black coat walks down the sidewalk holding a red umbrella".}\label{RedUmbrella}
\end{subfigure}\\
\begin{subfigure}[t]{3.4cm}
\includegraphics[width=1\linewidth]{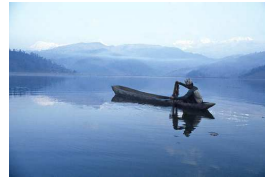}
\caption{``A man in a boat on a lake with mountains in the background".}\label{Boat}
\end{subfigure}\qquad
\begin{subfigure}[t]{3.6cm}
\includegraphics[width=1\linewidth]{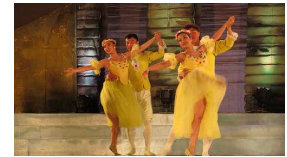}
\caption{``Four performers are performing with their arms outstretched in a ballet".}\label{FourPerformers}
\end{subfigure}
\caption{Examples of generated image captions using the proposed SuperCaptioning method.\label{CaptionExamples}}
\end{figure}

In this paper, we address the image captioning problem by employing the two-dimensional word embedding from the Super Characters method, and the resulting method is named as SuperCaptioning method. In this method, the input image and the raw text are combined together through two-dimensional embedding, and then fed into a CNN model to sequentially predict the words in the image caption. The experimental results on Flickr30k shows that the proposed method gives high quality image captions. Some examples given by SuperCaptioning method are shown in Figure~\ref{CaptionExamples}.

\begin{figure*}[t]
\begin{center}
   \includegraphics[width=1\linewidth]{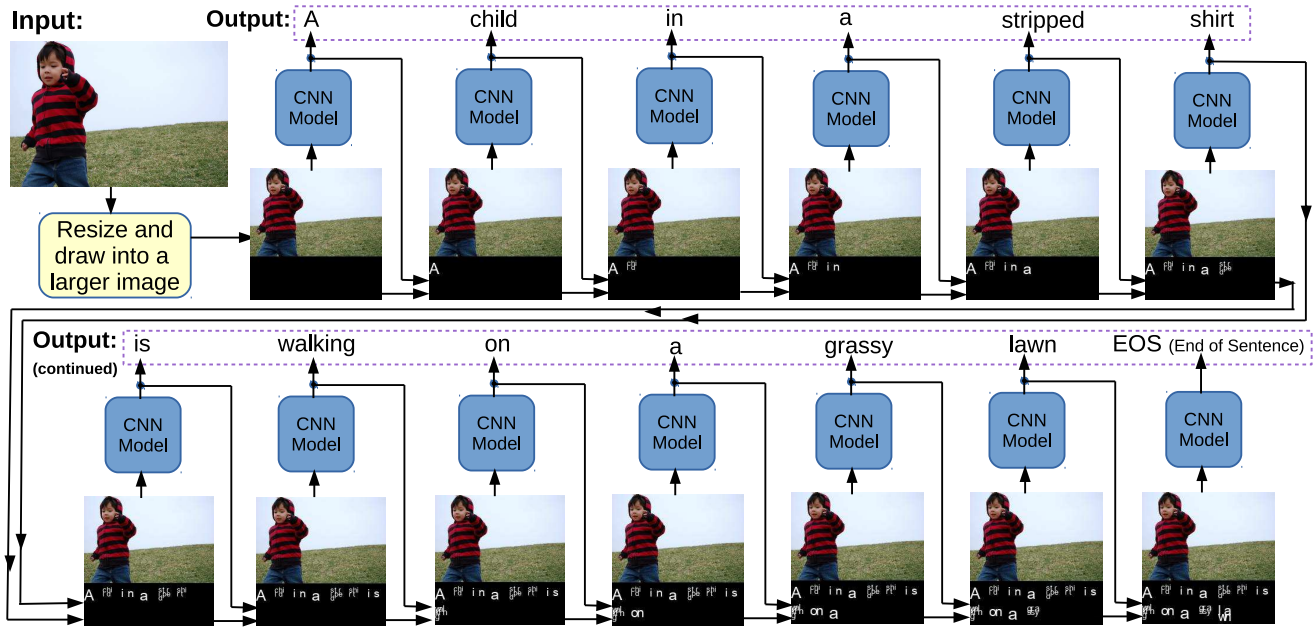}
\end{center}
   \caption{SuperCaptioning method illustration. The output caption is ``A child in a striped shirt is walking on a grassy lawn".}
\label{SuperCaptioningFigure}
\end{figure*}
\section{The Proposed SuperCaptioning Method}
The SuperCaptioning method is motivated by the success of Super Characters method on text classification tasks. Super Characters method converts text into images. So it will be very natural to combine the input image and the image of the text together, and feed it into one single CNN model to predict the next word in the image caption sentence.

Figure~\ref{SuperCaptioningFigure} illustrates the proposed SuperCaptioning method. The caption is predicted sequentially by predicting the next word in multiple iterations. At the beginning of the caption prediction, the partial caption is initialized as null, and the input image is resized to occupying a designed portion (e.g. top) of a larger blank image as shown in Figure~\ref{SuperCaptioningFigure}. Then the text of the current partial caption is drawn into the the other portion (e.g. bottom) of the larger image as well. The resulting image is called the SuperCaptioning image, which is then fed into a CNN model to classify the next word in the caption. The CNN model is fine-tuned from the ImageNet pre-trained model. The iteration continues until the next word is EOS (End Of Sentence) or the word count reaches the cut-length. Cut-length is defined as the maximum number of words for the caption.

Squared English Word (SEW) method is used to represent the English word in a squared space. For example, the word ``child" occupies the same size of space as the word ``A", but each of its alphabet will only occupies $\{1/ceil[sqrt(N)]\}^2$ of the word space, where $N$ is five for ``child" which has five alphabets, $sqrt(.)$ stands for square root, and $ceil[.]$ is rounding to the top.

The data used for training is from Flickr30k\footnote{http://shannon.cs.illinois.edu/DenotationGraph/data/flickr30k.html}. Each image in Flickr30k has 5 captions by different people, and we only keep the longest caption if it is less than 14 words as the ground truth caption for the training data. After this filtering, 31,333 of the total 31,783 images are left. 

After comparing the accuracy of experimental results using different configurations for the font size, cut-length, and resizing of the input image, we finally set the font size to 31, cut-length to 14 words, and resizing the image size to 150x224 in the fixed-size SuperCaptioning image with 224x224 pixels, as shown in Figure~\ref{SuperCaptioningFigure}.

The training data is generated by labeling each SuperCaptioning image as an example of the class indicated by its next caption word. EOS is labeled to the SuperCaptioning image if the response sentence is finished. The model used is SE-net-154~\cite{hu2018squeeze} pre-trained on ImageNet\footnote{https://github.com/hujie-frank/SENet}. We fine-tuned this model on our generated data set by only modifying the last layer to 11571 classes, which is the vocabulary size of all the selected captions.

Figure~\ref{CaptionExamples} shows that the proposed SuperCaptioning method captions the number of objects in the image, as shown in Figure~\ref{FourOrange} ``{\bf Four men} ..."; and it also captions the colors of overlapping objects, as shown in Figure~\ref{RedUmbrella} ``A woman in a {\bf black coat} ... holding a {\bf red umbrella}"; it captions the big picture of the background, as shown in Figure~\ref{Boat} ``... with {\bf mountains in the background}"; and it also captions the detailed activity, as shown in Figure~\ref{FourPerformers} ``... with their {\bf arms outstretched} in a ballet".

\section{Conclusion}
In this paper, we propose the SuperCaptioning method for image captioning using two-dimensional word embedding. The experimental results on Flickr30k shows that the SuperCaptioning method gives high quality image captions. The proposed method could be used for on-device image captioning applications with low-power CNN accelerator becoming more and more available~\cite{sun2018ultra,sun2018mram}. An interactive demo is ready to show at the workshop. 
{\small
\bibliographystyle{ieee_fullname}
\bibliography{egbib}
}

\end{document}